\documentclass[journal]{journal}
\usepackage{authblk}

\ifCLASSINFOpdf
\usepackage[pdftex]{graphicx}
\else
\fi

\usepackage[cmex10]{amsmath}
\let\labelindent\relax
\usepackage{enumitem}

\usepackage{mdwmath}
\usepackage{mdwtab}

\begin{document}

%
\title{Scalable Pooled Time Series of Big Video Data from the Deep Web}
%
%
%

\author[1,2]{Chris A. Mattmann}
\author[1]{Madhav Sharan}
\affil[1]{University of Southern California, Los Angeles, CA 90089 USA \authorcr Email: {\tt \{msharan,mattmann\}@usc.edu}\vspace{1.5ex}}
\affil[2]{Jet Propulsion Laboratory, California Institute of Technology, Pasadena, CA 91109 USA \authorcr Email: {\tt chris.a.mattmann@jpl.nasa.gov} \vspace{1.5ex}} 

\markboth{Journal of \LaTeX\ Class Files,~Vol.~6, No.~1, January~2007}%
{Shell \MakeLowercase{\textit{et al.}}: Scalable Pooled Time Series of Big Video Data from the Deep Web}

\maketitle
\thispagestyle{empty}

\begin{abstract}
We contribute a scalable implementation of Ryoo et al's Pooled Time Series algorithm from CVPR 2015. The updated algorithm has been evaluated on a large and diverse dataset of approximately 6800 videos collected from a crawl of the deep web related to human trafficking on DARPA's MEMEX effort. We describe the properties of Pooled Time Series and the motivation for using it to relate videos collected from the deep web. We highlight issues that we found while running Pooled Time Series on larger datasets and discuss solutions for those issues. Our solution centers are re-imagining Pooled Time Series as a Hadoop-based algorithm in which we compute portions of the eventual solution in parallel on large commodity clusters. We demonstrate that our new Hadoop-based algorithm works well on the 6800 video dataset and shares all of the properties described in the CVPR 2015 paper. We suggest avenues of future work in the project.
\end{abstract}


\IEEEpeerreviewmaketitle

\section{Introduction}

\IEEEPARstart{M}{}ultimedia content available today on the web is increasingly diverse, dynamic, and important. Besides online media such as YouTube collecting 24 hours of video/minute or \textasciitilde158Tb data/minute\footnote{Using a simple heuristic of 4.7Gb/DVD size per 2 hours of video.} \cite{davidson2010youtube} for use in social interactions, instructional videos, personal and family sharing of daily moments, etc., reliance on video has largely increased in other domains. For example, images and video are used in online product sales such as e-commerce of regular consumer products.

We have seen an increase in video content used in other commerce markets which we call ``dark markets'' that refer to illegal trade and sale of goods. In particular, working on the DARPA MEMEX \cite{fox2015memex} effort our team has worked in conjunction with law enforcement to mine multimedia content such as videos on the web to help thwart human trafficking; weapons and arms trafficking, and trafficking of other products, such as counterfeit electronics. In these dark markets which are present on the deep web -- that is the web behind \texttt{forms, and logins}, behind \texttt{javascript and Ajax interactions} and the web behind \texttt{heterogeneous content} -- increased use of multimedia content is a tool used by those selling their wares to remain hidden from traditional web searches and from bulk analysis. Through DARPA MEMEX our team and dozens of other teams are working to augment search engine technologies to automatically enable bulk analysis even on multimedia types.

We have collected is a 6800+ video set culled from a series of sites on the deep web on which human trafficking occurs. The dataset is 26Gb in size and 6805 in number (including duplicates); 14.3 Gb and 3266 in number (without duplicates). All of the videos are in MP4 format with an average video size of 3.8MB, and at least 1Mb, and at most \textasciitilde10Mb in the deduplicated set - in the set with duplicates the average size is 4Mb, the minimum size 1Mb and the maximum size  \textasciitilde10Mb. We will refer to this dataset as \texttt{HT video dataset} throughout the rest of the paper. 

One of the core research activities that we have explored on this dataset is whether or not we could reliably and at scale draw relationships between the videos therein. We used the Apache Tika system \cite{mattmann2011tika} initially for our analysis. Tika is a content detection and analysis system that automatically identifies a file's type; and automatically selects a relevant parser to extract text, metadata and language information from that file. Tika has support for Multimedia metadata extraction such as EXIF metadata that is present in images and videos and that tells us scene, content editing, and authorship properties about the video. As shown in our prior work on multimedia metadata forensics \cite{murdock2016second} - simply looking at the metadata is fast, reliable, and provides context relating similarly edited images and videos, and even sometimes relates multimedia content with similar scene properties. 

However, often on the web, the multimedia metadata is not enough and actual analysis of images and video must be done in order to find actual pixel or {\em gradient} based relationships. With video we also have the ability to use {\em temporal} relationships that show us properties such as {\em motion} present in the videos and so forth. 

Building upon our team's prior work in this area, we investigated using the Pooled Time Series (PoT) approach \cite{ryoo2015pot} that combines both oriented gradient (image differencing) and optical flow (image motion differencing) to approximate how related videos are based on those descriptors. PoT provides a reliable, and accurate relationship between multimedia videos using the aforementioned two key descriptors. It is efficient since it provides a temporal summary of those descriptors over configurable time intervals and has been shown to be effective in grouping and relating first-person videos, such as those in on the ground robotics, in defense related exercises for soldiers, and/or in situations where video must be collected by autonomous devices due to treacherous terrain. 

Our hypothesis was that the PoT algorithm would also be useful in videos from the \texttt{HT video dataset}. Based on early inspection many of them had these types of properties related to trafficking victims. Relating videos together could help identify relationships between victims; common scenes and housing features ({\em gradient}); common movements such as dancing and/or derobing ({\em flow}); common clothing ({\em gradient}) and other relationships that could better enable law enforcement to aid these victims.

In early applications of PoT to smaller datasets, we were able to successfully execute it on video sizes of 100, and 500 videos within reasonable amounts of time (maximum: 2 days). However, as soon as we entered the realm of 1000, 5000 and eventually \textasciitilde7000 videos, we were never able to get the PoT reference implementation (coded in Java) to complete on the dataset. We found this was due to various reasons in particular that we will outline below:
\setlist[description]{leftmargin=\parindent,labelindent=\parindent}
\begin{description}
\item[Out of Memory (OoM) issues] Since the PoT code computes time-interval based descriptions and must pool and summarize those descriptor vectors for motion and for gradient difference over many video frames over many 1000s of videos, we constantly ran out of memory even on a 32Gb machine in Amazon's cloud.
\item[Sequential Code] The steps of PoT - which we will outline in this paper - were sequentially implemented in Java. That is, first the histogram of oriented gradients (HoG) was computed; then the histogram of optical flow (HoF) is computed - each of these are very large matrices with N dimensional vectors stored for each video. Then the differences between HoG and HoF for each video are computed. Then the Mean Chi Square is computed summarizing the differences.
\item[Instrumentation and Checkpointing] The code lacked the necessary logging information to indicate where and when it broke. Since the code was initially implemented procedurally as a series of functions called by a main driver in Java it was difficult to trace the control flow and to figure out where to log errors.
\end{description}

While we were able to address the {\em Instrumentation and Checkpointing} issues in our existing Java code, addressing OoM issues and sequential code prevented us from computing the PoT's HoG and HoF descriptors on our \texttt{HT video dataset}. Since our team includes a member of the Apache Nutch committee that helped to build Apache Hadoop \cite{white2012hadoop}, we were inspired to see if PoT was an algorithm that we could develop a Hadoop-version of that would address the OoM and sequential code scalability problems that we were running in to. Hadoop is an open source implementation of Google's Map Reduce \cite{dean2008mapreduce} and Google File System (GFS) \cite{ghemawat2003google} for fast parallel data processing and highly reliable, available and redundant data on commodity clusters. Hadoop versions of algorithms regularly run in minutes compared to days, hours or months as they make effective use of the underlying commodity hardware. A version of PoT on Hadoop, to our knowledge, has never been implemented.

Creating a Hadoop version of an algorithm that previously ran sequentially typically involves: (1) defining a function to split a large input dataset (in this case sets of videos from \texttt{HT video dataset}); (2) defining a function a {\em Mapper} to compute on individually split smaller versions of the dataset in parallel; (3) defining a function a {\em Reducer} that combines the partial output results from the mappers and creates final output dataset(s) resulting from the processing. We will describe our process of applying this methodology to create a Hadoop version of PoT or \texttt{Hadoop-POT} for short in the remainder of the paper. The process will be evaluated with respect to its ability to allow completion of PoT computation on the \texttt{HT video dataset}, and also will be qualitatively explained and evaluated.

A roadmap for the paper follows. Sec.~\ref{sec:pot-approach} describes the general steps of PoT motivating their conversion to Hadoop.  Sec.~\ref{sec:hadoop-pot} describes the steps of our \texttt{Hadoop-POT} algorithm and approach. We evaluate our approach in Sec.~\ref{sec:evaluation}. Sec.~\ref{sec:conclusion} rounds out the paper.

\section{The Pooled Time Series Approach for Video Similarity} 
\label{sec:pot-approach}
Pooled Time Series (PoT) is an algorithm that takes as input a set of $N$ videos and generates as output an $N x N$ matrix with the pair-wise similarities between each video $1...N$. The diagonals of the matrix are unused, as it half of the matrix since it contains duplicative information. The first step in computing PoT is generating two histograms: one histogram of oriented gradients (HoG) to identify object movements and one histogram of optical flow (HoF) to model relative motion between actor and viewer in a video. Together these features make our PoT representation of a video. Further subsections provide more detail on the PoT process.

\subsection{Generate time series for each video}
For each video $1...N$ we generate HoF and HoG time series representing each frame of a video. To compute these histograms, we must preprocess each video. Each original file is resized and then changed into gray scale for easier processing.

\subsection{HoF time series generation}
HoF computation starts by creating a three dimensional $5$x$5$x$8$ array. Optical flow is evaluated using the Gunnar Farneback function \cite{farneback2000fast}. The algorithm takes frames of a video and computes the neighborhood of both frames as quadratic polynomials, and then computes global displacement. Global displacement is calculated by examining the yields from the quadratic polynomials and trying to equate the coefficients from those yields. This is demonstrated in the following equation.
\begin{equation}
prev(y,x) \approx next(y + flow(y,x)[1],x + flow(y,x)[0] )
\end{equation}
The above equation compares the previous image to the next one from the second frame onwards. Once the flow is calculated, the $x$ magnitude, the $y$ magnitude and orientation of the each pixel in the file is calculated. The optical flow is stored in the appropriate bin in the three dimensional $5$x$5$x$8$ array. A histogram is then created using the flows.

\subsection{HoG time series generation} 
The HoG computation also starts by creating a three dimensional $5$x$5$x$8$ array. We interpolate the differences between pixels in consecutive frames using the bilinear interpolation method as shown below:
\begin{equation}
f(x,y)=(A(1-x)+Bx)(1-y)+(C(1-x)+Dx)y
\end{equation}

The range of the difference between $x$ and $y$ pixels is 0 or 255, and based on the actual difference value, we force the values to be one of those two choices. The interpolation is done for each of the 8 orientations possible and finally we create a histogram using the gradient values resulting from this computation.

\subsection{Computation of PoT feature vector}
We compute the final PoT feature vector using both of the above series and by applying temporal filters and three pooling operators: $p_{i}$ sum pooling - $p_1$, gradient pooling - $p_2$, max pooling - $p_3$.

\subsection{Calculation of the raw similarity scores} 
To calculate raw similarity scores we generate all possible pairs of video combinations. If initial video size is $N$ we end up $\frac{N(N-1)}{2}$ pairs. We will denote these pairs using $N \times N$ matrix $V$ where $V_{ij}$ is pair of video $v_i$ and video $v_j$ and $i,j \in [1,N]$. For each pair of videos we assign a similarity score.

\subsection{Mean chi squared distance}
In PoT, we then calculate the chi squared distance (CSD) between all the features of all the possible video pairs. This gives us the distance number between each pair of videos. For each video we have two series histograms (HoF and HoG) and three pooling operators so for each video we have total of six features. We will denote each video feature of video $v_i \in i [1-N]$, sth pooling series $s \in [1,2]$ pth pooling operator $p \in [1,3]$ as $fv_{i,s,p}$. Here $fv_{i,s,p}$ is a vector of size call it $M$ which is obtained for each video, series and pooling operation. For each pair $V_{ij},s, p$ we calculate 
\begin{equation}
CSD_{i,j,s,p} = \frac{1}{2} \sum_{x}^{M} \frac{ (fv_{i,s,p,x} - fv_{j,s,p,x} )^2} { fv_{i,s,p,x} + fv_{j,s,p,x} }
\end{equation}

Then we calculate mean chi squared distance (\texttt{MEAN\_CSD}) as:
\begin{equation}
\texttt{MEAN\_CSD}_{s,p} = \sum_{i=1}^{N} \sum_{j=1}^{N} \frac{CSD_{i,j,s,p}  }{ \frac{N(N-1)}{2} }
\end{equation}

The combination of s and p can give us six values so we use these values to compute the mean chi squared distance for each s and p.

\subsection{Kernel distance}
Once we calculate mean chi squared distance for whole video set we calculate the all kernel distances (KD) by calculating chi square distance for a given series and pooling operator between two videos. We then divide the chi square distance by its corresponding mean chi square distance. Adding all the kernel distances for one video pair gives us the total distance between the pair. This is demonstrated in the following equation:

\begin{equation}
KD(i,j) = \sum_{s=1}^{2} \sum_{p=1}^{3} (CSD_{i,j,s,p} / \texttt{MEAN\_CSD}_{s,p})
\end{equation}

We calculate final PoT ``score'' (representative of the {\em similarity} between two videos) by passing x = KD(i,j) in below function:
\begin{equation}
f(x) = exp(\frac{-x}{10})
\end{equation}

We use this function for the following reasons. First for any positive value of $x$, the maximum value of $f(x)$ is 1 and in our case $x$ will always be positive as it is calculated through the chi squared distance. Second as the value of $x$ increases the value of $f(x)$ also decreases. Finally, the value of $f(x)$ is never negative. In summary the above function converts distances in the range $[0,\infty ]$ to a similarity score ranging $[1,0]$ making the scores more manageable, easier to plot and compare, and visualize, etc.

\subsection{Time complexity of the PoT algorithm} 
The overall time complexity of the PoT algorithm is roughly $O(N^2)$. The main factor that influences complexity is the number of videos, all other factors like number of series and pooling operators are constant and chosen manually. We were able to run this algorithm on a small set of $20-30$ videos easily but as $N$ increases the PoT performance decreases substantially. For \texttt{HT video dataset} there are nearly $24.5$ million pairs. If both pairs have $800$ frames with a $200$ pixel representation in HoF and HoG we are processing $640,000$ raw values for each pair. This restricts PoT to a few hundred videos to complete in any meaningful amount of time - we ran it for nearly a week without completion on a 20 node Amazon cloud cluster, where each node was M3.xlarge, with 32Gb of RAM running CentOS Linux. We manually split the video files up and ran them in small sequential steps representing different portions of the PoT algorithm e.g., HoF and HoG generation; computation of raw similarity scores, mean chi squared distance and kernel distance.

In the ensuing section we will detail our Hadoop implementation of this algorithm which is the core contribution of our paper.

\section{Hadoop Pooled Time Series}
\label{sec:hadoop-pot} 
Re-envisioning the PoT algorithm as a Hadoop Map Reduce job was an iterative process. In considering the core representation of the large dataset to split, we settled on the {\em list of video files} as the core data structure which would be split and partitioned using Hadoop Map Reduce. We will briefly provide some background on Hadoop-based processing. For a detailed treatment, see \cite{white2012hadoop} or many of the other myriad resources available that describe both the implementation and theory behind Hadoop.

\subsection{Some Early Considerations}
For our purposes, using Hadoop requires us to define the following constructs: (1) a {\em Split/Partition} function that will partition the video datasets into equally sized, independent jobs that can proceed in parallel; (2) a {\em Mapper} function, that, given a set of equally sized independent data from the split, will process that data, and produce some intermediate results; (3) a {\em Reducer} function, that takes the intermediate output from the {\em Mapper} jobs, and that combines those intermediate results from several {\em Mapper}s into final output results. In defining these functionality, we configure the Hadoop framework. Hadoop handles {\em Mapper} and {\em Reducer} job execution; re-execution in the case of failure; data input and output storage in a highly redundant and available filesystem, HDFS, and final output storage and results. Hadoop provides out of the box capabilities to split datasets (via hash; other uniqueness functions, and so forth), so there was not direct need to define an input split, we simply used what was available from Hadoop.

We considered that since the PoT is actually a data-flow dependent algorithm in the sense that e.g., HoF and HoG must be calculated {\em before} raw similarity score generation, that the raw similarity scores need to be computed {\em before} mean chi squared computation, and that mean chi square values must be present {\em before} the final kernel distance can be computed, that we would be unable to define a single Mapper and Reducer job to represent each of these steps. Instead, the best way to proceed would be to define several sets of Map/Reduce combination tasks, representing each of the stages of the algorithm that could individually be ran in parallel speeding up each step and ultimately the computation of the final result. The next section lists some of our early considerations for constructing the algorithm.

\subsection{Implementation Considerations}
For the actual implementation of the algorithm, we considered using the Hadoop Image Processing Interface (HIPI) \cite{sweeney2011hipi}, but instead chose to use core OpenCV since the early implementation of the sequential PoT algorithm was done using direct calls to OpenCV and we wanted to maintain as much backwards compatibility with the prior code as possible. We also looked at the Xuggle API \cite{lan2011p2p} but its main focus was on streaming video and our MP4 videos from the \texttt{HT video dataset} were not streaming.

We realized early on that we could have separate Mapper/Reducer tasks for computation of HoG and HoF, and that those largely could be done once, offline, and then kept in HDFS for future use in the other Mapper/Reducer task combinations. For example, once the HoF and HoG were generated for all videos, we could create a Reducer that could be used to compute similarities (initially raw), and also mean chi squared distance and kernel distance. 

We also desired to keep the output format for the matrix simple, so we chose a simple ASCII-text file to represent the output $N$x$N$ matrix to remain backwards compatible with the research code we had, and also to take advantage of Hadoop HDFS and its native support for text files. The PoT Java framework provided mechanisms to easily convert this matrix to a JSON file if needed. HoF and HoG outputs were similarly stored as matrices in ASCII oriented text file formats (such as CSV), as were the output raw similarity scores and distances. Though some of our thinking in this area would evolve (as we will detail later), those were our initial considerations. 

The final design we arrived at for Hadoop PoT includes two jobs for video processing of HoF and HoG, one job for calculating mean distance (using the raw distances and mean chi squared) and another job for calculating similarity (using kernel distance).

\subsection{Each Hadoop job of \texttt{Hadoop-POT}}
The specific Hadoop jobs and their architecture are explained in the ensuing sections.

\subsubsection{Video processing jobs} 
We have two jobs one each for calculating HoF and HoG from videos. Each of these jobs takes a file path as parameter and uses OpenCV to calculate features. These features are saved as \textit{video\_name.of.txt} and \textit{video\_name.hog.txt}. After completion of these two jobs we have two files for each video representing optical flow and optical gradient. Complexity is $O(N)$ as each video is processed only once.

\subsubsection{Similarity calculation} 
Similarity calculation requires all pairs of videos so the first step is to generate a cartesian product of video set with itself. Consider that we are processing a video set of three videos. For this, we will generate pair like $v_1v_2, v_1v_3, v_2v_3$. These pair of video names are stored in a CSV file which is provided as input to both similarity calculating jobs. In the above case we will have a file with three records. The first job calculates mean chi square distance for whole data set. It reads \textit{*.of.txt} and \textit{*.hog.txt} for each respective video pair and generates one output file having mean chi squared distance. We calculate mean chi squared distance for each series and for each pooling operator which gives us $2 \times 3 = 6$ mean distances. Our {\em Mapper} jobs in this step calculate chi squared distances for one pair and the associated {\em Reducer} job calculates the  mean of all the distances. We create a second {\em Mapper} job that takes a pair of filenames as input and that then calculates similarity score using the kernel distance. The output of this job is a CSV file with a similarity score assigned to each video.

\subsubsection{\texttt{Hadoop-POT} v1 and its Initial Flaws} 
The original \texttt{Hadoop-POT} v1 design was created by dumping the output results (CSVs, text files) of individual steps from HDFS to the local file system and then by copying those local file system results from the local file system to HDFS for the next stage of \texttt{Hadoop-POT}. This was performed for the following reason - we initially had great difficulty getting the Java OpenCV APIs to read data from HDFS. Errors ranged from intermittent exceptions, to full up I/O errors with Java itself. We later traced these to the implementation bindings in the Java OpenCV library, but were unable to find time as of yet to submit fixes to the API to make it compatible with Hadoop. 

Another key design issue with our v1 implementation was that while dumping \textit{video\_name.of.txt} and \textit{video\_name.hog.txt} we saved \texttt{double} values in a text file. This caused both of our similarity jobs to read these files  repeatedly $O(N^2)$ times and to, during the process, convert them from \texttt{string} values to \texttt{double} values for each value in the matrices. Considering previous example {\em Mapper} jobs had to convert $640,000$ values to \texttt{double} at a cost of about 1.5 seconds per file, scaling this up to 24 million jobs as required by \texttt{HT video dataset} would require more than a year of processing time which we did not have. While examining this issue we also realized that we were calculating features repeatedly for each pair although they were the same every time and could be saved.

Considering data formats as input to each jobs also forced us to question using individual output files for each of the videos rather than simply using native Hadoop constructs like \texttt{SequenceFiles} which can represent matrices and vectors and store {\em all of the values} for our video data set in a single file. Our initial approach of using a CSV with file paths pointing to the computed text files for HoF and HoG for each video masked the real input and input size was thus determined via a single file, rather than the actual total $N$ video files present in the dataset. This led to a smaller input splits as Hadoop had no idea about the size of the real input. Input splits should contain values of features rather than path to features for fair input splits. Using less input splits caused resources on the cluster in which we were evaluating v1 of our algorithm to be underutilized and as a result, our \texttt{Hadoop-POT} v1 was quite slow. 

Apart from input splits with this input structure we could not use Hadoop's distributed file processing effectively. Hadoop HDFS is made to deal with large files wherein which a file is broken into small splits and each split can be on a different machine. When a job is assigned to a container it is assigned to container that is running on same machine where the split is stored to reduce IO and contention. With input files being paths to features we were unable to use Hadoop's I/O data locality, and often our {\em Mapper} jobs were reading data from different nodes instead of those that contained the actual data part of the split.

\subsection{Intermediate Experiments with Redis - \texttt{Hadoop-POT} v2} 
Our v2 \texttt{Hadoop-POT} architecture addressed the v1 initial design flaws in the following ways. To save time in converting \texttt{text} to \texttt{double} format for the values in our matrices we setup a Redis cache server. Redis is an efficient, scalable key value based caching utility that operates in a client server fashion with automatic data typing. A client connects to a Redis server and then \texttt{PUTS} a key, value pair, where the key can be any typed data, and so can the value. Redis takes the responsibility of efficiently storing that key, value pair, providing a \texttt{GET} based interface to retrieve it later rapidly and efficiently. We built our v2 system using a Redis cache with the key being a single video file name and the value being computed feature vector. This brought down our job time down to milliseconds instead of 1.5 seconds per computation as in the v1 architecture. However, in doing so, we introduced yet another component into our architecture besides Hadoop and went around Hadoop's native support for similar data structures.

We also tried resource localization by archiving all the \textit{video\_name.of.txt} and \textit{video\_name.hog.txt} into one big zip file and passing that file as a cached archive in our Hadoop similarity jobs. This helped in reducing time spent at reading the file for the very first time. We also experimented with different split sizes to increase resource utilization. All of these experiments helped in improving the above mentioned flaws but again increased the overall complexity of our system in that it introduced a new system and framework that was not Hadoop.

\subsection{\texttt{Hadoop-POT}-v3} 
\begin{figure}
  \includegraphics[width=\linewidth]{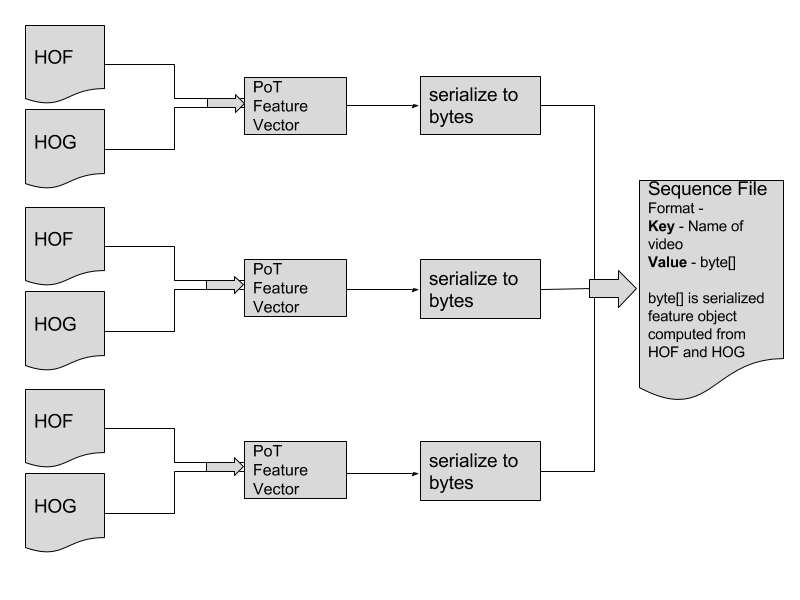}
  \caption{Sequence file generation using HoF and HoG features}
  \label{fig:SEQ_FILE}
\end{figure}

Realizing that Hadoop provided native data structures for fast storage and retrieval of multiple objects, and structures that themselves were easily splittable and usable as input to {\em Mapper} jobs, and {\em Reducer} jobs, we tried to eliminate the Redis component introduced in v2 of our architecture. \texttt{SequenceFile}s are a native Hadoop construct that allow easy combination and summarization of many small input files. Sequence files can be thought of as a zipped version of multiple files. As shown in Fig.~\ref{fig:SEQ_FILE} we take both features of one video and compute its PoT representation as a java object and serialize that object to bytes in v3 of our architecture. This serialized vector is stored as the value in sequence file, wherein which the key is the name of video file. We will use this key later while outputting the overall PoT similarity CSV as output of the algorithm. This sequence file is also generated natively by our {\em Mapper} jobs and read natively by our {\em Reducer} jobs, removing any specialized code that we had present in v2 of our architecture to talk to Redis and localizing our entire solution on top of the Hadoop architecture. 

\begin{figure}
  \includegraphics[width=\linewidth]{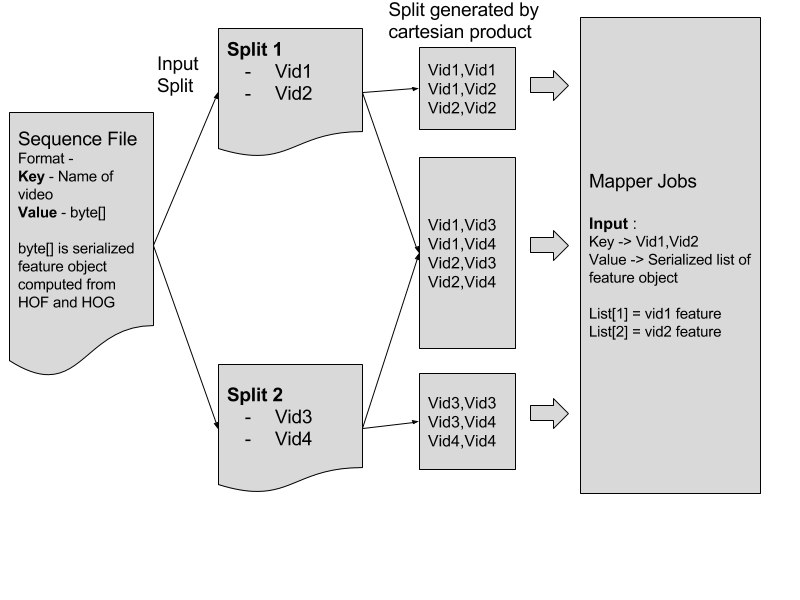}
  \caption{Cartesian product of sequence file with another sequence file generated from PoT.}
  \label{fig:cart}
\end{figure}

The final step in our v3 architecture was the need to implement computation of the Cartesian product on sequence files. We did this by overriding Hadoop's \texttt{RecordReader} and \texttt{FileInputFormat} classes. Fig.~\ref{fig:cart} explains how a \texttt{SequenceFile} is divided into splits and how video pairs are provided to {\em Mapper} jobs through \texttt{CartesianInputFormat}. Our similarity jobs take generated \texttt{SequenceFile}s as input and use our own \texttt{CartesianInputFormat} which first permutes through splits and then permutes within a split to generate a pair of video features for our {\em Mapper jobs}.

\subsection{Performance}
Our \texttt{Hadoop-POT} v3 algorithm reduced the time necessary for each of our {\em Mapper} jobs to complete on average to around 42 milliseconds evaluated on \texttt{HT video dataset} allowing the needed 24 million+ computations to be executable in a reasonable amount of time. This entire \texttt{HT video dataset} was processed using \texttt{Hadoop-POT} v3 on a 10 node cluster in Amazon using M3.xlarge instances with 32Gb memory each, and with EBS storage with 8 containers on each node in $\approx$25 hours. We used cached vector files for this run which takes $\approx$8 hours to generate from input videos in the \texttt{HT video dataset}.

\section{Qualitative Evaluation}
\label{sec:evaluation}
Qualitatively evaluating our results involves multiple areas. First and foremost, simply being able to compute PoT on the \texttt{HT video dataset} which we were unable to do using the previous PoT algorithm and implementation is the main contribution of our work. Without our \texttt{Hadoop-POT} v3 algorithm and its implementation, large video datasets cannot be run through the PoT algorithm and implementation.

\begin{figure}
  \includegraphics[width=\linewidth]{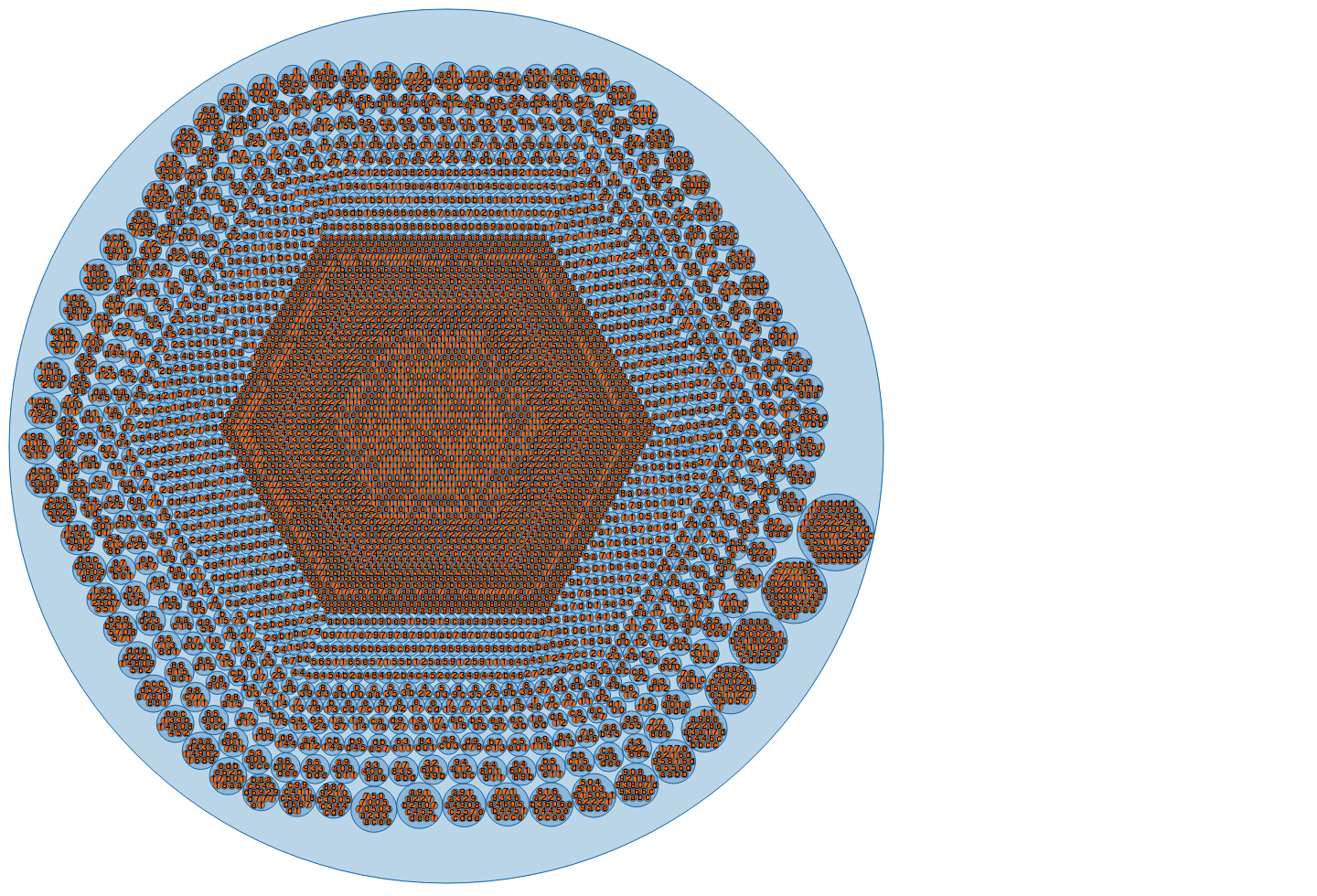}
  \caption{Clusters capturing deduplication in the \texttt{HT video dataset}}
  \label{fig:circlepacking}
\end{figure}

Considering now the actual results of running \texttt{Hadoop-POT} on the \texttt{HT video dataset} and considering whether PoT is a valid metric for those videos presents some interesting discussion points. One of the satisfying results was identifying almost identical pairs. These pair of videos had same content but they differed in their hash codes due to subtle differences and hence would not be caught through hashing. These results can be seen through Fig.~\ref{fig:circlepacking} The same video clips of different length (One video being subset of other) also received a high similarity score e.g., $0.9$. Results were also very satisfying in comparison of videos with different resolutions. For example if we compare the same video in $800 \times 600$ format with its own self in $1024 \times 768$ we will get a similarity score of near about $1$. 

\begin{figure}
  \includegraphics[width=\linewidth]{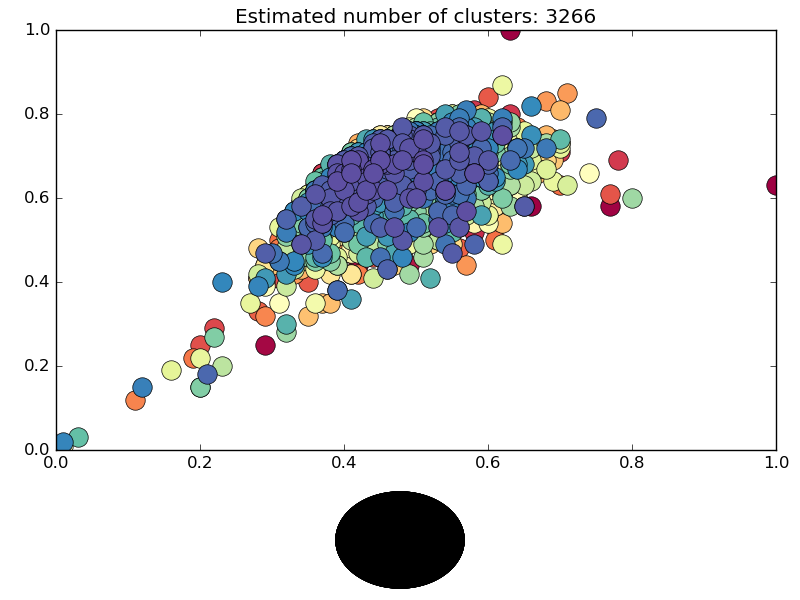}
  \caption{Near similar clusters for the deduplicated \texttt{HT video dataset}}
  \label{fig:similarity_cluster_21}
\end{figure}
\begin{figure}
  \includegraphics[width=\linewidth]{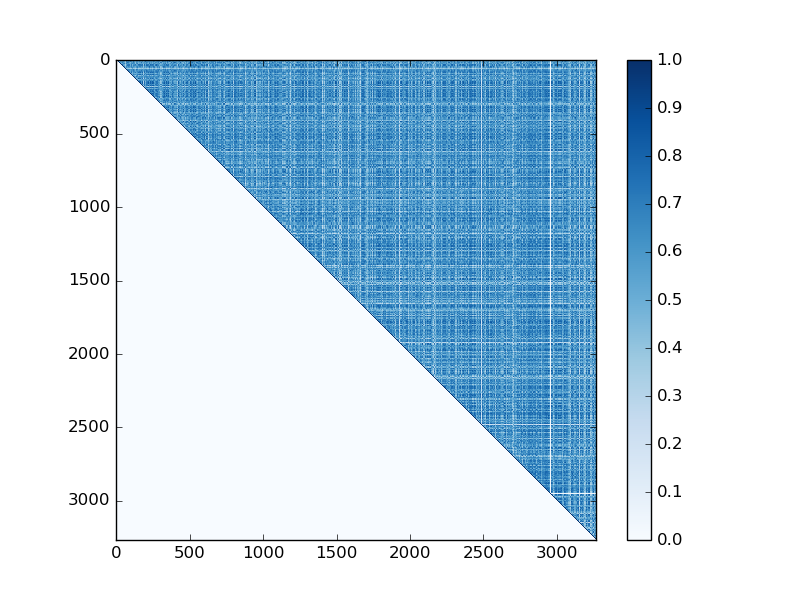}
  \caption{Heat map showing related videos from \texttt{Hadoop-POT} from the \texttt{HT video dataset}}
  \label{fig:similarity_heatmap_21}
\end{figure}

Another benefit of \texttt{Hadoop-POT} involved comparing videos with similar motion events and with similar backgrounds e.g., clusters having similar videos shown in Fig.~\ref{fig:similarity_cluster_21} and in Fig.~\ref{fig:similarity_heatmap_21}. Many of the grouped videos from our dataset brought together videos with trafficking victims taking selfies, or holding a similar note with contact information. That PoT captured both similar objects, motion, and background in these sets was encouraging. Looking beyond human trafficking, we downloaded a handful ($<$20) of videos related to the London Olympics and Usain bolt, and were able to apply \texttt{Hadoop-POT} v3 to those videos as well. We compared video of Usain bolt running in London olympics with video of Usain bolt running in Rio olympics and found similar groups, similarity metrics, and cluster results that were captured by PoT. Having the ability to run PoT on large input datasets collected from the (deep) web or otherwise, is a direct enabler of its application to large datasets.

Looking critically at the results, we also found some areas for improvements. We saw some videos in which a person did similar hand and head movements but that that were not found by PoT or grouped as similar or related. Our intuition in these examples are that HoF and HoG were unable to effectively capture certain movements, e.g., patterns or shapes,and as such adding more descriptors to the base implementation of PoT would likely aid these situations.

For future research we are planning to add few preprocessing steps including: (1) removing banners at starting of a video; and (2) dividing a video into a set of scenes. These steps will help in comparing videos of dissimilar length and removing unwanted parts from the video which may interfere with base feature comparisons.

\section{Conclusion}
\label{sec:conclusion}
The Pooled Time Series (PoT) algorithm and its ability to bring together video features for motion and for gradient differencing showing similarities and differences between motion and scene objects, people and places was of direct interest to our work in stopping human trafficking, arms trafficking, and other forms of dark markets on the deep web. When faced with the inability to generate a PoT as a video comparison metric and considering that PoT had shown promise in limited experiments on sets of 10-100-500 videos, we were challenged to re-envision the algorithm using scalable data processing and computational platform, namely Apache Hadoop.  

We went through several design considerations and prototyped three different version of the PoT algorithm as a series of Hadoop {\em Mapper} and {\em Reducer} jobs - the end result we call \texttt{Hadoop-POT}. We explain the tradeoffs in our design and what worked and what did not work. We point the reader to future areas of improvement and show how our re-envisioned PoT architecture was able to successfully complete on a large dataset from the human trafficking domain of 6800+ videos with interesting properties.

Our future work includes adding more pre-processing steps to \texttt{Hadoop-POT} and also making it easier for users to partition videos to isolate interesting features of motion and scenery.

\section*{Acknowledgment}
This work was supported by the DARPA XDATA/Memex program. In addition, the NSF Polar Cyberinfrastructure award numbers PLR-1348450 and PLR-144562 funded a portion of the work. Effort supported in part by JPL, managed by the California Institute of Technology on behalf of NASA.

The authors would like to thank Dr. Mattmann's DR student Soumya Ravi and Michael Joyce of NASA JPL for working with Dr. Michael Ryoo on hardening the initial research implementation of the pooled time series code for MEMEX. We also thank contributions from Dr. Ryoo on the original PoT implementation in Java.

\ifCLASSOPTIONcaptionsoff
  \newpage
\fi

\bibliographystyle{IEEEtran}
\bibliography{IEEEabrv,icmr2017} 

\end{document}